%% file: c3d_video.tex
\documentclass[10pt,twocolumn,letterpaper]{article}

\usepackage{iccv}
\usepackage{times}
\usepackage{epsfig}
\usepackage{graphicx}
\usepackage{amsmath}
\usepackage{amssymb}
\usepackage{color}

\usepackage[pagebackref=true,breaklinks=true,letterpaper=true,colorlinks,bookmarks=false]{hyperref}

\iccvfinalcopy 

\def\iccvPaperID{'625} 

\ificcvfinal\fi
\begin{document}

\title{Learning Spatiotemporal Features with 3D Convolutional Networks}

\author{Du Tran$^{1,2}$, Lubomir Bourdev$^1$, Rob Fergus$^1$, Lorenzo Torresani$^2$, Manohar Paluri$^1$\\
$^1$Facebook AI Research, $^2$Dartmouth College\\
{\tt\small \{dutran,lorenzo\}@cs.dartmouth.edu} \ \ \ {\tt\small \{lubomir,robfergus,mano\}@fb.com}
}

\maketitle

\vspace{-12pt}
\input{abstract}

\vspace{-12pt}
\input{introduction}

\vspace{-6pt}
\input{related_work}

\input{feature_learning}

\input{action_recognition}

\input{action_similarity_labeling}

\input{scene_object}

\input{efficiency}

\input{conclusions}

\input{resolution_study}

\input{learned_filters}

{\footnotesize
\bibliographystyle{ieee}
\bibliography{ieeedu_ref}
}

\end{document}

%% file: abstract.tex
\begin{abstract}

We propose a simple, yet effective approach for spatiotemporal feature learning using deep 3-dimensional convolutional networks (3D ConvNets) trained on a large scale supervised video dataset. Our findings are three-fold: 1) 3D ConvNets are more suitable for spatiotemporal feature learning compared to 2D ConvNets; 2) A homogeneous architecture with small $3 \times 3 \times 3$ convolution kernels in all layers is among the best performing architectures for 3D ConvNets; and 3) Our learned features, namely C3D (Convolutional 3D), with a simple linear classifier outperform state-of-the-art methods on 4 different benchmarks and are comparable with current best methods on the other 2 benchmarks. In addition, the features are compact: achieving $52.8\%$ accuracy on UCF101 dataset with only $10$ dimensions and also very efficient to compute due to the fast inference of ConvNets. Finally, they are conceptually very simple and easy to train and use.

\end{abstract}

%% file: introduction.tex
\section{Introduction}
\label{sec:intro}

Multimedia on the Internet is growing rapidly resulting in an increasing number of videos being shared every minute. To combat the information explosion it is essential to understand and analyze these videos for various purposes like search, recommendation, ranking etc. The computer vision community has been working on video analysis for decades and tackled different problems such as action recognition~\cite{Laptev03}, abnormal event detection~\cite{Boiman07}, and activity understanding~\cite{Kitani12}. Considerable progress has been made in these individual problems by employing different specific solutions. However, there is still a growing need for a generic video descriptor that helps in solving large-scale video tasks in a homogeneous way.

\begin{table*}
\begin{center}
{\small
\begin{tabular}{|l|c|c|c|c|c|c|}
\hline
Dataset & Sport1M & UCF101 & ASLAN & YUPENN & UMD & Object \\
Task & {\footnotesize action recognition} & {\footnotesize action recognition} & {\footnotesize action similarity labeling} & {\footnotesize scene classification} & {\footnotesize scene classification} & {\footnotesize object recognition} \\ 
\hline
Method & \cite{Ng15} & \cite{SrivastavaMS15}(\cite{LanLLHR14}) & \cite{PengWWQ14} & \cite{FeichtenhoferPW14} & \cite{FeichtenhoferPW14} & \cite{Ren2009}\\
Result & {\bf 90.8} & 75.8 (89.1) & 68.7 & 96.2 & 77.7 & 12.0 \\
{\bf C3D} & 85.2 & {\bf 85.2} ({\bf 90.4}) & {\bf 78.3} & {\bf 98.1} & {\bf 87.7} & {\bf 22.3}\\
\hline
\end{tabular}
}
\end{center}
\vspace{-6pt}
\caption{{\bf C3D compared to best published results}. C3D outperforms all previous best reported methods on a range of benchmarks except for Sports-1M and UCF101. On UCF101, we report accuracy for two groups of methods. The first set of methods use only RGB frame inputs while the second set of methods (in parentheses) use all possible features (e.g. optical flow, improved Dense Trajectory).} 
\label{tab:summary_result}
\vspace{-6pt}
\end{table*}

There are four properties for an effective video descriptor: (i) it needs to be {\bf generic}, so that it can represent different types of videos well while being discriminative. For example, Internet videos can be of landscapes, natural scenes, sports, TV shows, movies, pets, food and so on; (ii) the descriptor needs to be {\bf compact}: as we are working with millions of videos, a compact descriptor helps processing, storing, and retrieving tasks much more scalable; (iii) it needs to be {\bf efficient} to compute, as thousands of videos are expected to be processed every minute in real world systems; and (iv) it must be {\bf simple} to implement. Instead of using complicated feature encoding methods and classifiers, a good descriptor should work well even with a simple model (e.g. linear classifier).

Inspired by the deep learning breakthroughs in the image domain~\cite{Krizhevsky12} where rapid progress has been made in the past few years in feature learning, various pre-trained convolutional network (ConvNet) models~\cite{jia2014caffe} are made available for extracting image features. These features are the activations of the network's last few fully-connected layers which perform well on transfer learning tasks~\cite{Ning14,Zhou2014}. However, such image based deep features are not directly suitable for videos due to lack of motion modeling (as shown in our experiments in sections~\ref{sec:action_recognition},\ref{sec:aslan},\ref{sec:dynamic_scene}). In this paper we propose to learn spatio-temporal features using deep $3$D ConvNet. We empirically show that these learned features with a simple linear classifier can yield good performance on various video analysis tasks. Although $3$D ConvNets were proposed before~\cite{Ming2013,Karpathy14}, to our knowledge this work exploits 3D ConvNets in the context of large-scale supervised training datasets and modern deep architectures to achieve the best performance on different types of video analysis tasks. The features from these 3D ConvNets encapsulate information related to objects, scenes and actions in a video, making them useful for various tasks without requiring to finetune the model for each task. C3D has the properties that a good descriptor should have: it is generic, compact, simple and efficient. To summarize, our contributions in this paper are:

\begin{itemize}
\item We experimentally show 3D convolutional deep networks are good feature learning machines that model appearance and motion simultaneously.
\vspace{-5pt}
\item We empirically find that $3 \times 3 \times 3$ convolution kernel for all layers to work best among the limited set of explored architectures.
\vspace{-5pt}
\item The proposed features with a simple linear model outperform or approach the current best methods on {\bf 4} different tasks and {\bf 6} different benchmarks (see Table~\ref{tab:summary_result}). They are also compact and efficient to compute.
\end{itemize}

%% file: related_work.tex
\section{Related Work}
\label{sec:relwork}

Videos have been studied by the computer vision community for decades. Over the years various problems like action recognition~\cite{Laptev03}, anomaly detection~\cite{Boiman07}, video retrieval~\cite{bendersky14}, event and action detection~\cite{TRECVID,THUMOS14}, and many more have been proposed. Considerable portion of these works are about video representations. Laptev and Lindeberg~\cite{Laptev03} proposed spatio-temporal interest points (STIPs) by extending Harris corner detectors to $3$D. SIFT and HOG are also extended into SIFT-$3$D~\cite{Scovanner07} and HOG3D~\cite{KMS08} for action recognition. Dollar~\emph{et al.} proposed Cuboids features for behavior recognition~\cite{Piotr05}. Sadanand and Corso built ActionBank for action recognition~\cite{ActionBank}. Recently, Wang~\emph{et al.} proposed improved Dense Trajectories (iDT)~\cite{Wang2013} which is currently the state-of-the-art hand-crafted feature. The iDT descriptor is an interesting example showing that temporal signals could be handled differently from that of spatial signal. Instead of extending Harris corner detector into 3D, it starts with densely-sampled feature points in video frames and uses optical flows to track them. For each tracker corner different hand-crafted features are extracted along the trajectory. Despite its good performance, this method is computationally intensive and becomes intractable on large-scale datasets.

With recent availability of powerful parallel machines (GPUs, CPU clusters), together with large amounts of training data, convolutional neural networks (ConvNets)~\cite{LeNet5} have made a come back providing breakthroughs on visual recognition~\cite{girshick13,Krizhevsky12}. ConvNets have also been applied to the problem of human pose estimation in both images~\cite{jain14iclr} and videos~\cite{jain14accv}. More interestingly these deep networks are used for image feature learning~\cite{Donahue13}. Similarly, Zhou~\emph{et al.} and perform well on transferred learning tasks. Deep learning has also been applied to video feature learning in an unsupervised setting~\cite{Le2011}. In Le~\emph{et al.}~\cite{Le2011}, the authors use stacked ISA to learn spatio-temporal features for videos. Although this method showed good results on action recognition, it is still computationally intensive at training and hard to scale up for testing on large datasets. $3$D ConvNets were proposed for human action recognition~\cite{Ming2013} and for medical image segmentation~\cite{jain10,turaga2010}. 3D convolution was also used with Restricted Boltzmann Machines to learn spatiotemporal features~\cite{taylor10}. Recently, Karpathy~\emph{et al.}~\cite{Karpathy14} trained deep networks on a large video dataset for video classification. Simonyan and Zisserman~\cite{SimonyanZ14} used two stream networks to achieve best results on action recognition. 

\begin{figure*}[t]
\begin{center}
   \includegraphics[width=0.9\linewidth]{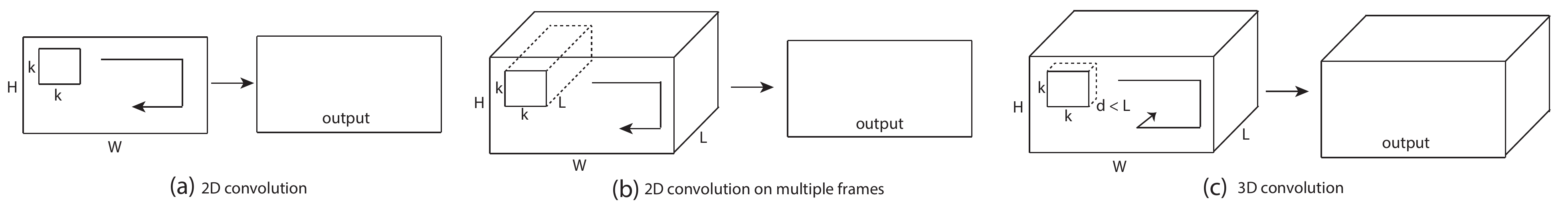}
\end{center}
\vspace{-12pt}
   \caption{{\bf 2D and 3D convolution operations}. a) Applying $2$D convolution on an image results in an image. b) Applying $2$D convolution on a video volume (multiple frames as multiple channels) also results in an image. c) Applying $3$D convolution on a video volume results in another volume, preserving temporal information of the input signal.}
\label{fig:conv3d_vs_conv2d}
\vspace{-12pt}
\end{figure*}

Among these approaches, the $3$D ConvNets approach in~\cite{Ming2013} is most closely related to us. This method used a human detector and head tracking to segment human subjects in videos. The segmented video volumes are used as inputs for a 3-convolution-layer 3D ConvNet to classify actions. In contrast, our method takes full video frames as inputs and does not rely on any preprocessing, thus easily scaling to large datasets. We also share some similarities with Karpathy~\emph{et al.}~\cite{Karpathy14} and Simonyan and Zisserman~\cite{SimonyanZ14} in terms of using full frames for training the ConvNet. However, these methods are built on using only 2D convolution and 2D pooling operations (except for the Slow Fusion model in~\cite{Karpathy14}) whereas our model performs 3D convolutions and 3D pooling propagating temporal information across all the layers in the network (further detailed in section~\ref{sec:learning_features}). We also show that gradually pooling space and time information and building deeper networks achieves best results and we discuss more about the architecture search in section~\ref{sec:arch_search}.

%% file: feature_learning.tex
\section{Learning Features with 3D ConvNets}
\label{sec:learning_features}
In this section we explain in detail the basic operations of 3D ConvNets, analyze different architectures for 3D ConvNets empirically, and elaborate how to train them on large-scale datasets for feature learning. 

\subsection{3D convolution and pooling}
We believe that $3$D ConvNet is well-suited for spatiotemporal feature learning. Compared to $2$D ConvNet, $3$D ConvNet has the ability to model temporal information better owing to $3$D convolution and $3$D pooling operations. In $3$D ConvNets, convolution and pooling operations are performed spatio-temporally while in 2D ConvNets they are done only spatially. Figure~\ref{fig:conv3d_vs_conv2d} illustrates the difference, $2$D convolution applied on an image will output an image, $2$D convolution applied on multiple images (treating them as different channels~\cite{SimonyanZ14}) also results in an image. Hence, $2$D ConvNets lose temporal information of the input signal right after every convolution operation. Only $3$D convolution preserves the temporal information of the input signals resulting in an output volume. The same phenomena is applicable for $2$D and $3$D polling.  In~\cite{SimonyanZ14}, although the temporal stream network takes multiple frames as input, because of the $2$D convolutions, after the first convolution layer, temporal information is collapsed completely. Similarly, fusion models in~\cite{Karpathy14} used $2$D convolutions, most of the networks lose their input's temporal signal after the first convolution layer. Only the \emph{Slow Fusion} model in~\cite{Karpathy14} uses $3$D convolutions and averaging pooling in its first $3$ convolution layers. We believe this is the key reason why it performs best among all networks studied in~\cite{Karpathy14}. However, it still loses all temporal information after the third convolution layer.

In this section, we empirically try to identify a good architecture for 3D ConvNets. Because training deep networks on large-scale video datasets is very time-consuming, we first experiment with UCF101, a medium-scale dataset, to search for the best architecture. We verify the findings on a large scale dataset with a smaller number of network experiments. According to the findings in 2D ConvNet~\cite{SimonyanZ14a}, small receptive fields of $3 \times 3$ convolution kernels with deeper architectures yield best results. Hence, for our architecture search study we fix the spatial receptive field to $3 \times 3$ and vary only the temporal depth of the 3D convolution kernels.

{\bf Notations}: For simplicity, from now on we refer video clips with a size of $c \times l \times h \times w$ where $c$ is the number of channels, $l$ is length in number of frames, $h$ and $w$ are the height and width of the frame, respectively. We also refer 3D convolution and pooling kernel size by $d \times k \times k$, where $d$ is kernel temporal depth and $k$ is kernel spatial size. 

{\bf Common network settings}: In this section we describe the network settings that are common to all the networks we trained. The networks are set up to take video clips as inputs and predict the class labels which belong to $101$ different actions. All video frames are resized into $128 \times 171$. This is roughly half resolution of the UCF101 frames. Videos are split into non-overlapped 16-frame clips which are then used as input to the networks. The input dimensions are $3 \times 16 \times 128 \times 171$. We also use jittering by using random crops with a size of $3 \times 16 \times 112 \times 112$ of the input clips during training. The networks have $5$ convolution layers and $5$ pooling layers (each convolution layer is immediately followed by a pooling layer), $2$ fully-connected layers and a softmax loss layer to predict action labels. The number of filters for $5$ convolution layers from $1$ to $5$ are $64$, $128$, $256$, $256$, $256$, respectively. All convolution kernels have a size of $d \time 3 \time 3$ where $d$ is the kernel temporal depth (we will later vary the value $d$ of these layers to search for a good $3$D architecture). All of these convolution layers are applied with appropriate padding (both spatial and temporal) and stride $1$, thus there is no change in term of size from the input to the output of these convolution layers. All pooling layers are max pooling with kernel size $2 \times 2 \times 2$ (except for the first layer) with stride $1$ which means the size of output signal is reduced by a factor of $8$ compared with the input signal. The first pooling layer has kernel size $1 \times 2 \times 2$ with the intention of not to merge the temporal signal too early and also to satisfy the clip length of $16$ frames (e.g. we can temporally pool with factor $2$ at most $4$ times before completely collapsing the temporal signal). The two fully connected layers have $2048$ outputs. We train the networks from scratch using mini-batches of $30$ clips, with initial learning rate of $0.003$. The learning rate is divided by $10$ after every $4$ epochs. The training is stopped after $16$ epochs.

{\bf Varying network architectures}: For the purposes of this study we are mainly interested in how to aggregate temporal information through the deep networks. To search for a good 3D ConvNet architecture, we only vary kernel temporal depth $d_i$ of the convolution layers while keeping all other common settings fixed as stated above. We experiment with two types of architectures: 1) homogeneous temporal depth: all convolution layers have the same kernel temporal depth; and 2) varying temporal depth: kernel temporal depth is changing across the layers. For homogeneous setting, we experiment with $4$ networks having kernel temporal depth of $d$ equal to $1$, $3$, $5$, and $7$. We name these networks as {\bf depth-d}, where $d$ is their homogeneous temporal depth. Note that \emph{depth}-$1$ net is equivalent to applying $2$D convolutions on separate frames. For the varying temporal depth setting, we experiment two networks with temporal depth {\bf increasing}: 3-3-5-5-7 and {\bf decreasing}: 7-5-5-3-3 from the first to the fifth convolution layer respectively. We note that all of these networks have the same size of the output signal at the last pooling layer, thus they have the same number of parameters for fully connected layers. Their number of parameters is only different at convolution layers due to different kernel temporal depth. These differences are quite minute compared to millions of parameters in the fully connected layers. For example, any two of the above nets with temporal depth difference of $2$, only has 17K parameters fewer or more from each other. The biggest difference in number of parameters is between \emph{depth}-$1$ net and \emph{depth}-$7$ net where \emph{depth}-$7$ net has 51K more parameters which is less than $0.3\%$ of the total of $17.5$ millions parameters of each network. This indicates that the learning capacity of the networks are comparable and the differences in number of parameters should not affect the results of our architecture search.

\subsection{Exploring kernel temporal depth}
\label{sec:arch_search}

\begin{figure}
\begin{center}
   \includegraphics[width=0.96\linewidth]{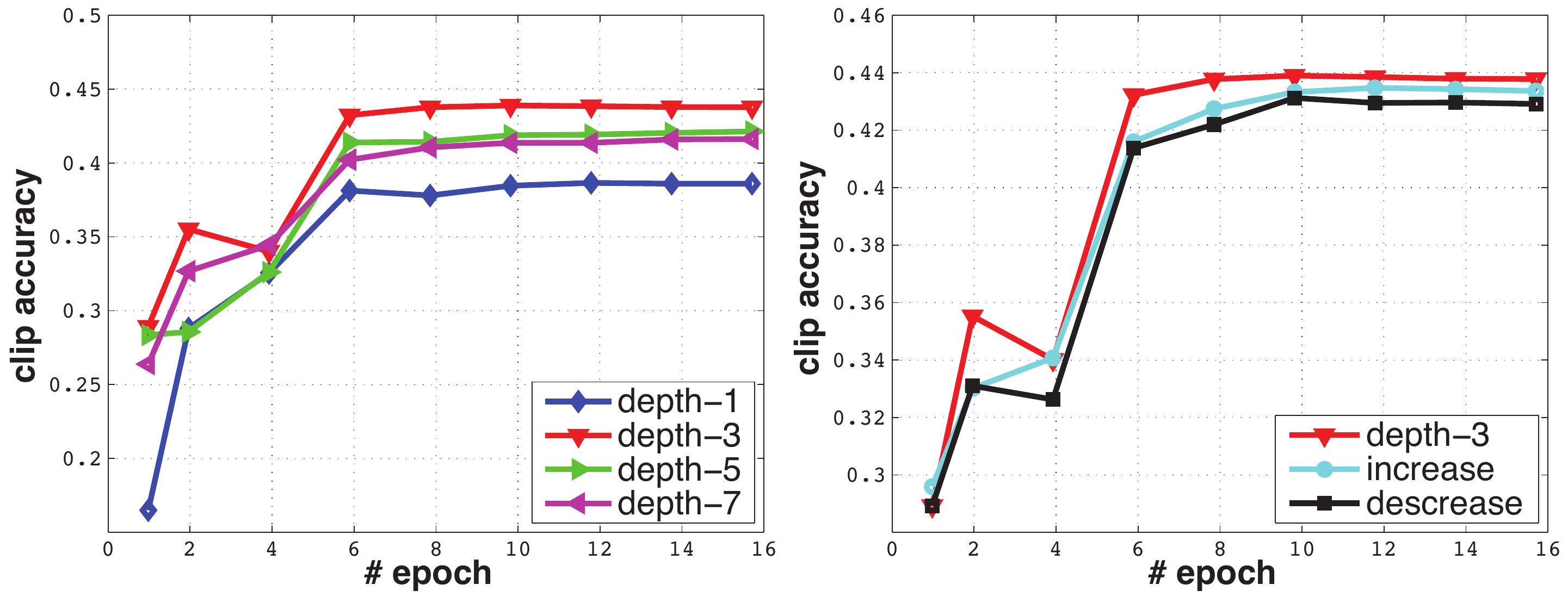}
\end{center}
\vspace{-12pt}
   \caption{{\bf 3D convolution kernel temporal depth search.} Action recognition clip accuracy on UCF101 test split-1 of different kernel temporal depth settings. $2$D ConvNet performs worst and $3$D ConvNet with $3 \times 3 \times 3$ kernels performs best among the experimented nets.}
\label{fig:c3d_arch_search}
\vspace{-12pt}
\end{figure}

We train these networks on the train split 1 of UCF101. Figure~\ref{fig:c3d_arch_search} presents clip accuracy of different architectures on UCF101 test split 1. The left plot shows results of nets with homogeneous temporal depth and the right plot presents results of nets that changing kernel temporal depth. \emph{Depth}-$3$ performs best among the homogeneous nets. Note that \emph{depth}-$1$ is significantly worse than the other nets which we believe is due to lack of motion modeling. Compared to the varying temporal depth nets, \emph{depth}-$3$ is the best performer, but the gap is smaller. We also experiment with bigger spatial receptive field (e.g. $5 \times 5$) and/or full input resolution ($240 \times 320$ frame inputs) and still observe similar behavior. This suggests $3 \times 3 \times 3$ is the best kernel choice for 3D ConvNets (according to our subset of experiments) and 3D ConvNets are consistently better than 2D ConvNets for video classification. We also verify that 3D ConvNet consistently performs better than 2D ConvNet on a large-scale internal dataset, namely I380K.


\begin{figure*}
\begin{center}
   \includegraphics[width=0.98\linewidth]{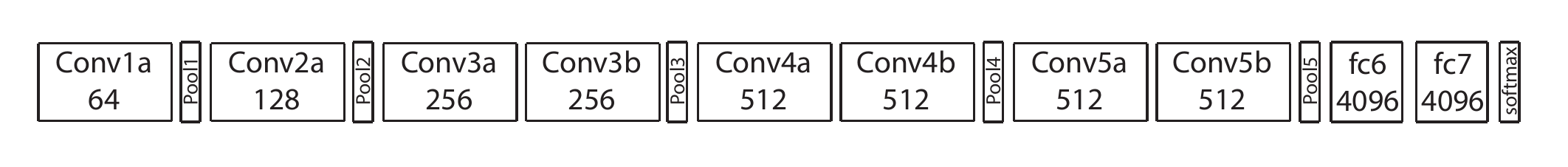}
\end{center}
\vspace{-18px}
   \caption{{\bf C$3$D architecture}. C3D net has $8$ convolution, $5$ max-pooling, and $2$ fully connected layers, followed by a softmax output layer. All $3$D convolution kernels are $3 \times 3 \times 3$ with stride $1$ in both spatial and temporal dimensions. Number of filters are denoted in each box. The $3$D pooling layers are denoted from \texttt{pool1} to \texttt{pool5}. All pooling kernels are $2 \times 2 \times 2$, except for \texttt{pool1} is $1 \times 2 \times 2$. Each fully connected layer has $4096$ output units.}
\label{fig:conv3d}
\end{figure*}

\begin{table*}
\begin{center}

\begin{tabular}{|l|c|c|c|c|}
\hline
{\bf Method} & {\bf Number of Nets} & {\bf Clip hit@1} & {\bf Video hit@1} & {\bf Video hit@5}\\ 
\hline
DeepVideo's Single-Frame + Multires~\cite{Karpathy14} & 3 nets & 42.4 & 60.0 & 78.5 \\
DeepVideo's Slow Fusion~\cite{Karpathy14} & 1 net & 41.9 & 60.9 & 80.2 \\
Convolution pooling on 120-frame clips~\cite{Ng15} & 3 net & {\bf 70.8}* & {\bf 72.4} & {\bf 90.8}\\
{\bf C3D} (trained from scratch) & 1 net & 44.9 & 60.0 & 84.4 \\
{\bf C3D} (fine-tuned from I380K pre-trained model) & 1 net & 46.1 & 61.1 & 85.2\\
\hline
\end{tabular}
\end{center}
\vspace{-6pt}
\caption{{\bf Sports-1M classification result}. C3D outperforms~\cite{Karpathy14} by $5\%$ on top-$5$ video-level accuracy. (*)We note that the method of \cite{Ng15} uses long clips, thus its clip-level accuracy is not directly comparable to that of C3D and DeepVideo.}
\label{tab:sport1m_result}
\vspace{-6pt}
\end{table*}

\subsection{Spatiotemporal feature learning}

{\bf Network architecture}: Our findings in the previous section indicate that homogeneous setting with convolution kernels of $3 \times 3 \times 3$ is the best option for 3D ConvNets. This finding is also consistent with a similar finding in 2D ConvNets~\cite{SimonyanZ14a}. With a large-scale dataset, one can train a 3D ConvNet with $3 \times 3 \times 3$ kernel as deep as possible subject to the machine memory limit and computation affordability. With current GPU memory, we design our 3D ConvNet to have 8 convolution layers, 5 pooling layers, followed by two fully connected layers, and a softmax output layer. The network architecture is presented in figure~\ref{fig:conv3d}. For simplicity, we call this net C3D from now on. All of $3$D convolution filters are $3 \times 3 \times 3$ with stride $1 \times 1 \times 1$. All $3$D pooling layers are $2 \times 2 \times 2$ with stride $2 \times 2 \times 2$ except for \texttt{pool1} which has kernel size of $1 \times 2 \times 2$ and stride $1 \times 2 \times 2$ with the intention of preserving the temporal information in the early phase. Each fully connected layer has $4096$ output units.

{\bf Dataset}. To learn spatiotemproal features, we train our C3D on Sports-1M dataset~\cite{Karpathy14} which is currently the largest video classification benchmark. The dataset consists of $1.1$ million sports videos. Each video belongs to one of $487$ sports categories. Compared with UCF101, Sports-1M has $5$ times the number of categories and $100$ times the number of videos.

{\bf Training}: Training is done on the Sports-1M train split. As Sports-1M has many long videos, we randomly extract five 2-second long clips from every training video. Clips are resized to have a frame size of $128 \times 171$. On training, we randomly crop input clips into $16 \times 112 \times 112$ crops for spatial and temporal jittering. We also horizontally flip them with $50\%$ probability. Training is done by SGD with mini-batch size of $30$ examples. Initial learning rate is $0.003$, and is divided by $2$ every 150K iterations. The optimization is stopped at 1.9M iterations (about $13$ epochs). Beside the C3D net trained from scratch, we also experiment with C3D net fine-tuned from the model pre-trained on I380K.

{\bf Sports-1M classification results}:  
Table~\ref{tab:sport1m_result} presents the results of our C3D networks compared with DeepVideo~\cite{Karpathy14} and Convolution pooling~\cite{Ng15}. We use only a single center crop per clip, and pass it through the network to make the clip prediction. For video predictions, we average clip predictions of $10$ clips which are randomly extracted from the video. It is worth noting some setting differences between the comparing methods. DeepVideo and C3D use short clips while Convolution pooling~\cite{Ng15} uses much longer clips. DeepVideo uses more crops: $4$ crops per clip and $80$ crops per video compared with $1$ and $10$ used by C3D, respectively. The C3D network trained from scratch yields an accuracy of $84.4\%$ and the one fine-tuned from the I380K pre-trained model yields $85.5\%$ at video top-$5$ accuracy. Both C3D networks outperform DeepVideo's networks. C3D is still $5.6\%$ below the method of~\cite{Ng15}. However, this method uses convolution pooling of deep image features on long clips of $120$ frames, thus it is not directly comparable to C3D and DeepVideo which operate on much shorter clips. We note that the difference in top-1 accuracy for clips and videos of this method is small ($1.6\%$) as it already uses 120-frame clips as inputs. In practice, convolution pooling or more sophisticated aggregation schemes~\cite{Ng15} can be applied on top of C3D features to improve video hit performance.

{\bf C3D video descriptor}: After training, C3D can be used as a feature extractor for other video analysis tasks. To extract C3D feature, a video is split into $16$ frame long clips with a 8-frame overlap between two consecutive clips. These clips are passed to the C3D network to extract \texttt{fc6} activations. These clip \texttt{fc6} activations are averaged to form a $4096$-dim video descriptor which is then followed by an L2-normalization. We refer to this representation as C3D video descriptor/feature in all experiments, unless we clearly specify the difference.

\begin{figure*}
\begin{center}
   \includegraphics[width=0.98\linewidth]{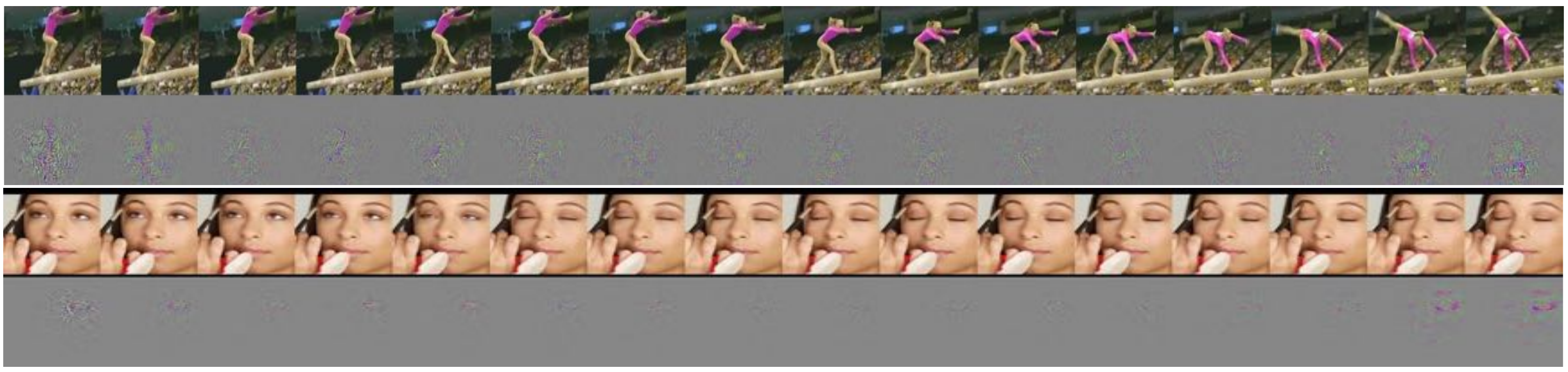}
\end{center}
\vspace{-12pt}
   \caption{{\bf Visualization of C3D model, using the method from~\cite{ZeilerF14}}. Interestingly, C3D captures appearance for the first few frames but thereafter only attends to salient motion. Best viewed on a color screen.}
\label{fig:deconv}
\end{figure*}

{\bf What does C3D learn?} We use the deconvolution method explained in~\cite{ZeilerF14} to understand what C3D is learning internally. We observe that C3D starts by focusing on appearance in the first few frames and tracks the salient motion in the subsequent frames. Figure~\ref{fig:deconv} visualizes deconvolution of two C3D \texttt{conv5b} feature maps with highest activations projected back to the image space. In the first example, the feature focuses on the whole person and then tracks the motion of the pole vault performance over the rest of the frames. Similarly in the second example it first focuses on the eyes and then tracks the motion happening around the eyes while applying the makeup. Thus C3D differs from standard 2D ConvNets in that it selectively attends to both motion and appearance. We provide more visualizations in the supplementary material to give a better insight about the learned feature. 

%% file: action_recognition.tex
\section{Action recognition}
\label{sec:action_recognition}
{\bf Dataset}: We evaluate C3D features on UCF101 dataset~\cite{UCF101}. The dataset consists of $13,320$ videos of $101$ human action categories. We use the three split setting provided with this dataset.

{\bf Classification model}: We extract C3D features and input them to a multi-class linear SVM for training models. We experiment with C3D descriptor using 3 different nets: C3D trained on I380K, C3D trained on Sports-1M, and C3D trained on I380K and fine-tuned on Sports-1M. In the multiple nets setting, we concatenate the L2-normalized C3D descriptors of these nets.

{\bf Baselines}: We compare C3D feature with a few baselines: the current best hand-crafted features, namely improved dense trajectories (iDT)~\cite{Wang2013} and the popular-used deep image features, namely Imagenet~\cite{jia2014caffe}, using Caffe's Imagenet pre-train model. For iDT, we use the bag-of-word representation with a codebook size of $5000$ for each feature channel of iDT which are trajectories, HOG, HOF, MBHx, and MBHy. We normalize histogram of each channel separately using L1-norm and concatenate these normalized histograms to form a 25K feature vector for a video. For Imagenet baseline, similar to C3D, we extract Imagenet \texttt{fc6} feature for each frame, average these frame features to make video descriptor. A multi-class linear SVM is also used for these two baselines for a fair comparison.

\begin{table}
\begin{center}

\begin{tabular}{|l|c|}
\hline
Method & Accuracy ($\%$) \\
\hline
Imagenet + linear SVM & 68.8 \\
iDT w/ BoW + linear SVM & 76.2 \\
\hline
Deep networks~\cite{Karpathy14} & 65.4 \\
Spatial stream network~\cite{SimonyanZ14} & 72.6 \\
LRCN~\cite{DonahueHGRVSD14} & 71.1 \\
LSTM composite model~\cite{SrivastavaMS15} & 75.8 \\
{\bf C3D} (1 net) + linear SVM & 82.3 \\
{\bf C3D} (3 nets) + linear SVM & {\bf 85.2} \\
\hline
iDT w/ Fisher vector~\cite{PengWWQ14} & 87.9\\
Temporal stream network~\cite{SimonyanZ14} & 83.7 \\
Two-stream networks~\cite{SimonyanZ14} & 88.0 \\
LRCN~\cite{DonahueHGRVSD14} & 82.9 \\
LSTM composite model~\cite{SrivastavaMS15} & 84.3 \\
Conv. pooling on long clips~\cite{Ng15} & 88.2 \\
LSTM on long clips~\cite{Ng15} & 88.6 \\
Multi-skip feature stacking~\cite{LanLLHR14} & 89.1 \\
{\bf C3D} (3 nets) + iDT + linear SVM& {\bf 90.4} \\
\hline
\end{tabular}
\end{center}
\vspace{-8pt}
\caption{{\bf Action recognition results on UCF101}. C3D compared with baselines and current state-of-the-art methods. Top: simple features with linear SVM; Middle: methods taking only RGB frames as inputs; Bottom: methods using multiple feature combinations.}
\label{tab:ucf101_result_combine}
\end{table}

{\bf Results}: Table~\ref{tab:ucf101_result_combine} presents action recognition accuracy of C3D compared with the two baselines and current best methods. The upper part shows results of the two baselines. The middle part presents methods that use only RGB frames as inputs. And the lower part reports all current best methods using all possible feature combinations (e.g. optical flows, iDT).

C3D fine-tuned net performs best among three C3D nets described previously. The performance gap between these three nets, however, is small ($1\%$). From now on, we refer to the fine-tuned net as C3D, unless otherwise stated. C3D using one net which has only $4,096$ dimensions obtains an accuracy of $82.3\%$. C3D with $3$ nets boosts the accuracy to $85.2\%$ with the dimension is increased to $12,288$. C3D when combined with iDT further improves the accuracy to $90.4\%$, while when it is combined with Imagenet, we observe only $0.6\%$ improvement. This indicates C3D can well capture both appearance and motion information, thus there is no benefit to combining with Imagenet which is an appearance based deep feature. On the other hand, it is beneficial to combine C3D with iDT as they are highly complementary to each other. In fact, iDT are hand-crafted features based on optical flow tracking and histograms of low-level gradients while C3D captures high level abstract/semantic information. 

\begin{figure}
\begin{center}
   \includegraphics[width=0.68\linewidth]{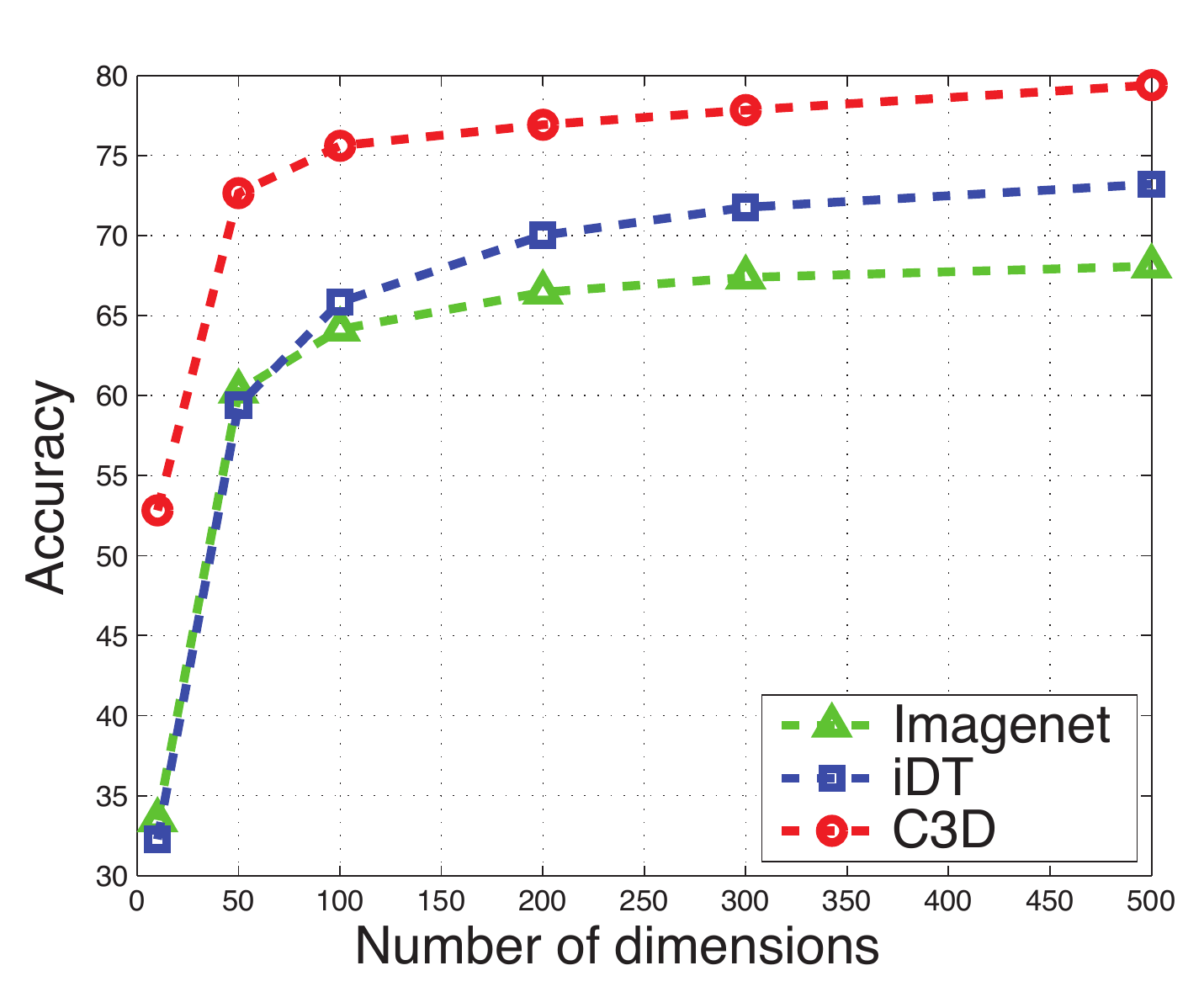}
\end{center}
\vspace{-12pt}
   \caption{{\bf C3D compared with Imagenet and iDT in low dimensions}. C3D, Imagenet, and iDT accuracy on UCF101 using PCA dimensionality reduction and a linear SVM. C3D outperforms Imagenet and iDT by $10$-$20\%$ in low dimensions.}
\label{fig:pca_ucf101}
\end{figure}

\begin{figure}[t]
\begin{center}
   \includegraphics[width=\linewidth]{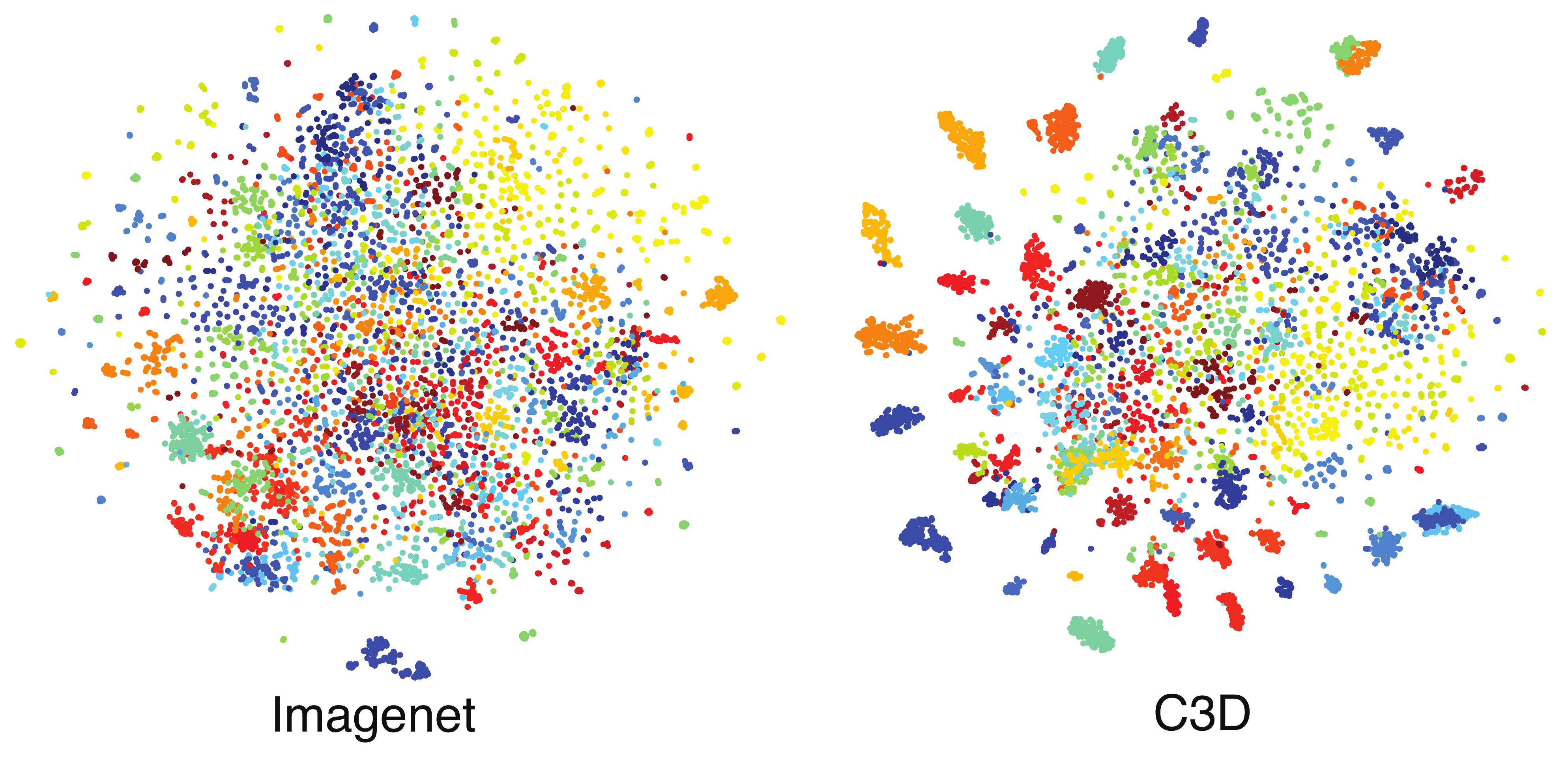}
\end{center}
\vspace{-12pt}
   \caption{{\bf Feature embedding}. Feature embedding visualizations of Imagenet and C3D on UCF101 dataset using t-SNE~\cite{van2008visualizing}. C3D features are semantically separable compared to Imagenet suggesting that it is a better feature for videos. Each clip is visualized as a point and clips belonging to the same action have the same color. Best viewed in color.}
\vspace{-16pt}
\label{fig:feature_embbeding}
\end{figure}

C3D with $3$ nets achieves $85.2\%$ which is $9\%$ and $16.4\%$ better than the iDT and Imagenet baselines, respectively. On the only RGB input setting, compared with CNN-based approaches, Our C3D outperforms deep networks~\cite{Karpathy14} and spatial stream network in~\cite{SimonyanZ14} by $19.8\%$ and $12.6\%$, respectively. Both deep networks~\cite{Karpathy14} and spatial stream network in~\cite{SimonyanZ14} use AlexNet architecture. While in~\cite{Karpathy14}, the net is fine-tuned from their model pre-trained on Sports-1M, spatial stream network in~\cite{SimonyanZ14} is fine-tuned from Imagenet pre-trained model. Our C3D is different from these CNN-base methods in term of network architecture and basic operations. In addition, C3D is trained on Sports-1M and used as is without any finetuning. Compared with Recurrent Neural Networks (RNN) based methods, C3D outperforms Long-term Recurrent Convolutional Networks (LRCN)~\cite{DonahueHGRVSD14} and LSTM composite model~\cite{SrivastavaMS15} by $14.1\%$ and $9.4\%$, respectively. C3D with only RGB input still outperforms these two RNN-based methods when they used both optical flows and RGB as well as the temporal stream network in~\cite{SimonyanZ14}. However, C3D needs to be combined with iDT to outperform two-stream networks~\cite{SimonyanZ14}, the other iDT-based methods~\cite{PengWWQ14,LanLLHR14}, and the method that focuses on long-term modeling~\cite{Ng15}. Apart from the promising numbers, C3D also has the advantage of simplicity compared to the other methods.

{\bf C3D is compact}: In order to evaluate the compactness of C3D features we use PCA to project the features into lower dimensions and report the classification accuracy of the projected features on UCF101~\cite{UCF101} using a linear SVM. We apply the same process with iDT~\cite{Wang2013} as well as Imagenet features~\cite{Donahue13} and compare the results in Figure~\ref{fig:pca_ucf101}. At the extreme setting with only $10$ dimensions, C3D accuracy is $52.8\%$ which is more than $20\%$ better than the accuracy of Imagenet and iDT which are about $32\%$. At $50$ and $100$ dim, C3D obtains an accuracy of $72.6\%$ and $75.6\%$ which are about $10$-$12\%$ better than Imagenet and iDT. Finally, with $500$ dimensions, C3D is able to achieve $79.4\%$ accuracy which is $6\%$ better than iDT and $11\%$ better than Imagenet. This indicates that our features are both compact and discriminative. This is very helpful for large-scale retrieval applications where low storage cost and fast retrieval are crucial.

We qualitatively evaluate our learned C3D features to verify if it is a good generic feature for video by visualizing the learned feature embedding on another dataset. We randomly select $100$K clips from UCF101, then extract \texttt{fc6} features for those clips using for features from Imagenet and C3D. These features are then projected to $2$-dimensional space using t-SNE~\cite{van2008visualizing}. Figure~\ref{fig:feature_embbeding} visualizes the feature embedding of the features from Imagenet and our C3D on UCF101. It is worth noting that we did not do any fine-tuning as we wanted to verify if the features show good generalization capability across datasets. We quantitatively observe that C3D is better than Imagenet.

%% file: action_similarity_labeling.tex
\section{Action Similarity Labeling}
\label{sec:aslan}
{\bf Dataset}: The ASLAN dataset consists of $3,631$ videos from $432$ action classes. The task is to predict if a given pair of videos belong to the same or different action. We use the prescribed $10$-fold cross validation with the splits provided with the dataset. This problem is different from action recognition, as the task focuses on predicting action similarity not the actual action label. The task is quite challenging because the test set contains videos of ``\emph{never-seen-before}'' actions.

\begin{figure}
\begin{center}
   \includegraphics[width=0.8\linewidth]{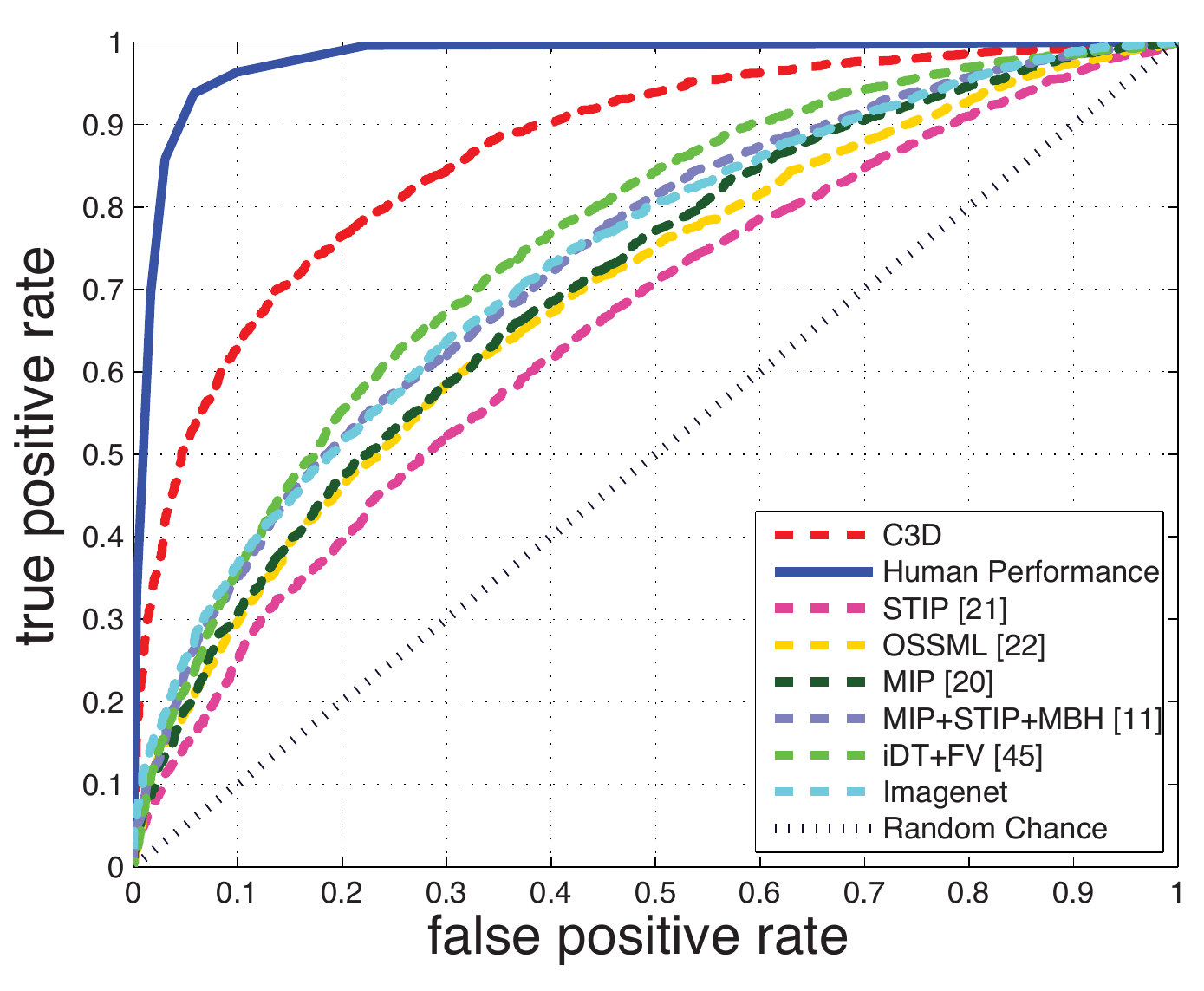}
\end{center}
\vspace{-12pt}
   \caption{{\bf Action similarity labeling result}. ROC curve of C3D evaluated on ASLAN. C3D achieves $86.5\%$ on AUC and outperforms current state-of-the-art by $11.1\%$.}
\label{fig:aslan_roc}
\end{figure}

{\bf Features}: We split videos into $16$-frame clips with an overlap of $8$ frames. We extract C3D features: \texttt{prob}, \texttt{fc7}, \texttt{fc6}, \texttt{pool5} for each clip. The features for videos are computed by averaging the clip features separately for each type of feature, followed by an L$2$ normalization.

{\bf Classification model}: We follow the same setup used in~\cite{aslanPAMI12}. Given a pair of videos, we compute the $12$ different distances provided in~\cite{aslanPAMI12}. With $4$ types of features, we obtain $48$-dimensional ($12 \times 4 = 48$) feature vector for each video pair. As these $48$ distances are not comparable to each other, we normalize them independently such that each dimension has zero mean and unit variance. Finally, a linear SVM is trained to classify video pairs into same or different on these $48$-dim feature vectors. Beside comparing with current methods, we also compare C3D with a strong baseline using deep image-based features. The baseline has the same setting as our C3D and we replace C3D features with Imagenet features. 

\begin{table}
\begin{center}
\begin{tabular}{|c|c|c|c|c|}
\hline
Method & Features & Model & Acc. & AUC \\
\hline
\cite{aslanPAMI12} & STIP & linear & 60.9 & 65.3 \\
\cite{aslanWS11} & STIP & metric & 64.3 & 69.1 \\
\cite{aslanECCV12} & MIP & metric & 65.5 & 71.9 \\
\cite{aslanCVPRW13} & MIP+STIP+MBH & metric & 66.1 & 73.2 \\
\cite{XPengSPL14} & iDT+FV & metric & 68.7 & 75.4 \\
Baseline & Imagenet & linear & 67.5 & 73.8 \\
{\bf Ours} & {\bf C3D} & linear & {\bf 78.3} & {\bf 86.5} \\ 
\hline
\end{tabular}
\end{center}
\vspace{-8pt}
\caption{{\bf Action similarity labeling result on ASLAN}. C3D significantly outperforms state-of-the-art method~\cite{XPengSPL14} by $9.6\%$ in accuracy and by $11.1\%$ in area under ROC curve.}
\label{tab:aslan_results}
\vspace{-16pt}
\end{table}

{\bf Results}: We report the result of C3D and compare with state-of-the-art methods in table~\ref{tab:aslan_results}. While most current methods use multiple hand-crafted features, strong encoding methods (VLAD, Fisher Vector), and complex learning models, our method uses a simple averaging of C3D features over the video and a \emph{linear} SVM. C3D significantly outperforms state-of-the-art method~\cite{XPengSPL14} by $9.6\%$ on accuracy and $11.1\%$ on area under ROC curve (AUC). Imagenet baseline performs reasonably well which is just $1.2\%$ below state-of-the-art method~\cite{XPengSPL14}, but $10.8\%$ worse than C3D due to lack of motion modeling. Figure~\ref{fig:aslan_roc} plots the ROC curves of C3D compared with current methods and human performance. C3D has clearly made a significant improvement which is a halfway from current state-of-the-art method to human performance ($98.9\%$).

%% file: scene_object.tex
\section{Scene and Object Recognition}
\label{sec:dynamic_scene}
\vspace{-6pt}
{\bf Datasets}: For dynamic scene recognition, we evaluate C3D on two benchmarks:  YUPENN~\cite{Derpanis12} and Maryland~\cite{umd_scene}. YUPENN consists of $420$ videos of $14$ scene categories and Maryland has $130$ videos of $13$ scene categories. For object recognition, we test C3D on egocentric dataset~\cite{Ren2009} which consists $42$ types of everyday objects. A point to note, this dataset is egocentric and all videos are recorded in a first person view which have quite different appearance and motion characteristics than any of the videos we have in the training dataset.

\begin{table}
\begin{center}
{\small
\begin{tabular}{|l|c|c|c|c|c|c|}
\hline
Dataset & \cite{Derpanis12} & \cite{Theriault13} &  \cite{FeichtenhoferPW13} & \cite{FeichtenhoferPW14} & Imagenet & {\bf C3D} \\
\hline
Maryland & 43.1 & 74.6 &  67.7 & 77.7 & {\bf 87.7} & {\bf 87.7} \\
YUPENN & 80.7 & 85.0 & 86.0 & 96.2 & 96.7 & {\bf 98.1} \\
\hline
\end{tabular}
}
\end{center}
\vspace{-8pt}
\caption{{\bf Scene recognition accuracy}. C3D using a simple linear SVM outperforms current methods on Maryland and YUPENN.}
\vspace{-16pt}
\label{tab:scene_soa}
\end{table}

{\bf Classification model}: For both datasets, we use the same setup of feature extraction and linear SVM for classification and follow the same leave-one-out evaluation protocol as described by the authors of these datasets. For object dataset, the standard evaluation is based on frames. However, C3D takes a video clip of length $16$ frames to extract the feature. We slide a window of $16$ frames over all videos to extract C3D features. We choose the ground truth label for each clip to be the most frequently occurring label of the clip. If the most frequent label in a clip occurs fewer than $8$ frames, we consider it as negative clip with no object and discard it in both training and testing. We train and test C3D features using linear SVM and report the object recognition accuracy. We follow the same split provided in~\cite{Ren2009}. We also compare C3D with a baseline using Imagenet feature on these 3 benchmarks. 

{\bf Results}: Table~\ref{tab:scene_soa} reports our C3D results and compares it with the current best methods. On scene classification, C3D outperforms state-of-the-art method~\cite{FeichtenhoferPW14} by $10\%$ and $1.9\%$ on Maryland and YUPENN respectively. It is worth nothing that C3D uses only a \emph{linear} SVM with simple averaging of clip features while the second best method~\cite{FeichtenhoferPW14} uses different \emph{complex} feature encodings (FV, LLC, and dynamic pooling). The Imagenet baseline achieves similar performance with C3D on Maryland and $1.4\%$ lower than C3D on YUPENN. On object recognition, C3D obtains $22.3\%$ accuracy and outperforms~\cite{Ren2009} by $10.3\%$ with only linear SVM where the comparing method used RBF-kernel on strong SIFT-RANSAC feature matching. Compared with Imagenet baseline, C3D is still $3.4\%$ worse. This can be explained by the fact that C3D uses smaller input resolution ($128 \times 128$) compared to full-size resolution ($256 \times 256$) using by Imagenet. Since C3D is trained only on Sports-1M videos without any fine-tuning while Imagenet is fully trained on $1000$ object categories, we did not expect C3D to work that well on this task. The result is very surprising and shows how generic C3D is on capturing appearance and motion information in videos.

%% file: efficiency.tex
\section{Runtime Analysis}
\label{sec:runtime}

We compare the runtime of C3D and with iDT~\cite{Wang2013} and the Temporal stream network~\cite{SimonyanZ14}. For iDT, we use the code kindly provided by the authors~\cite{Wang2013}. For~\cite{SimonyanZ14}, there is no public model available to evaluate. However, this method uses Brox's optical flows~\cite{BroxM11} as inputs. We manage to evaluate runtime of Brox's method using two different versions: CPU implementation provided by the authors~\cite{BroxM11} and the GPU implementation provided in OpenCV.

\begin{table}
\begin{center}
{\small
\begin{tabular}{|l|c|c|c|c|c|}
\hline
Method & iDT & Brox's & Brox's & C3D \\
Usage & CPU & CPU & GPU & GPU \\
\hline
RT (hours) & 202.2 & 2513.9 & 607.8 & 2.2 \\
FPS & 3.5 & 0.3 & 1.2 & 313.9 \\
x Slower & 91.4 & 1135.9 & 274.6 & 1 \\
\hline
\end{tabular}
}
\end{center}
\vspace{-8pt}
\caption{{\bf Runtime analysis on UCF101}. C3D is $91$x faster than improved dense trajectories~\cite{Wang2013} and $274$x faster than Brox's GPU implementation in OpenCV.}
\vspace{-16pt}
\label{tab:runtime}
\end{table}

We report runtime of the three above-mentioned methods to extract features (including I/O) for the whole UCF101 dataset in table~\ref{tab:runtime} using using a single CPU or a single K40 Tesla GPU. \cite{SimonyanZ14} reported a computation time (without I/O) of $0.06$s for a pair of images. In our experiment, Brox's GPU implementation takes $0.85$-$0.9$s per image pair including I/O. Note that this is not a fair comparison for iDT as it uses only CPU. We cannot find any GPU implementation of this method and it is not trivial to implement a parallel version of this algorithm on GPU. Note that C3D is much faster than real-time, processing at {\bf 313 fps} while the other two methods have a processing speed of less than $4$ fps.

%% file: conclusions.tex
\section{Conclusions}
\label{sec:conclusion}
In this work we try to address the problem of learning spatiotemporal features for videos using 3D ConvNets trained on large-scale video datasets. We conducted a systematic study to find the best temporal kernel length for 3D ConvNets. We showed that C3D can model appearance and motion information simultaneously and outperforms the 2D ConvNet features on various video analysis tasks. We demonstrated that C3D features with a linear classifier can outperform or approach current best methods on different video analysis benchmarks. Last but not least, the proposed C3D features are efficient, compact, and extremely simple to use.

C3D source code and pre-trained model are available at~\url{http://vlg.cs.dartmouth.edu/c3d}.

{\bf Acknowledgment}: we would like to thank Yann Lecun for his valuable feedback, Nikhil Johri and Engineering at Facebook AI Research for data and infrastructure support.

%% file: resolution_study.tex
\subsection*{Appendix A: Effects of Input Resolution}

\begin{figure}
\begin{center}
   \includegraphics[width=0.9\linewidth]{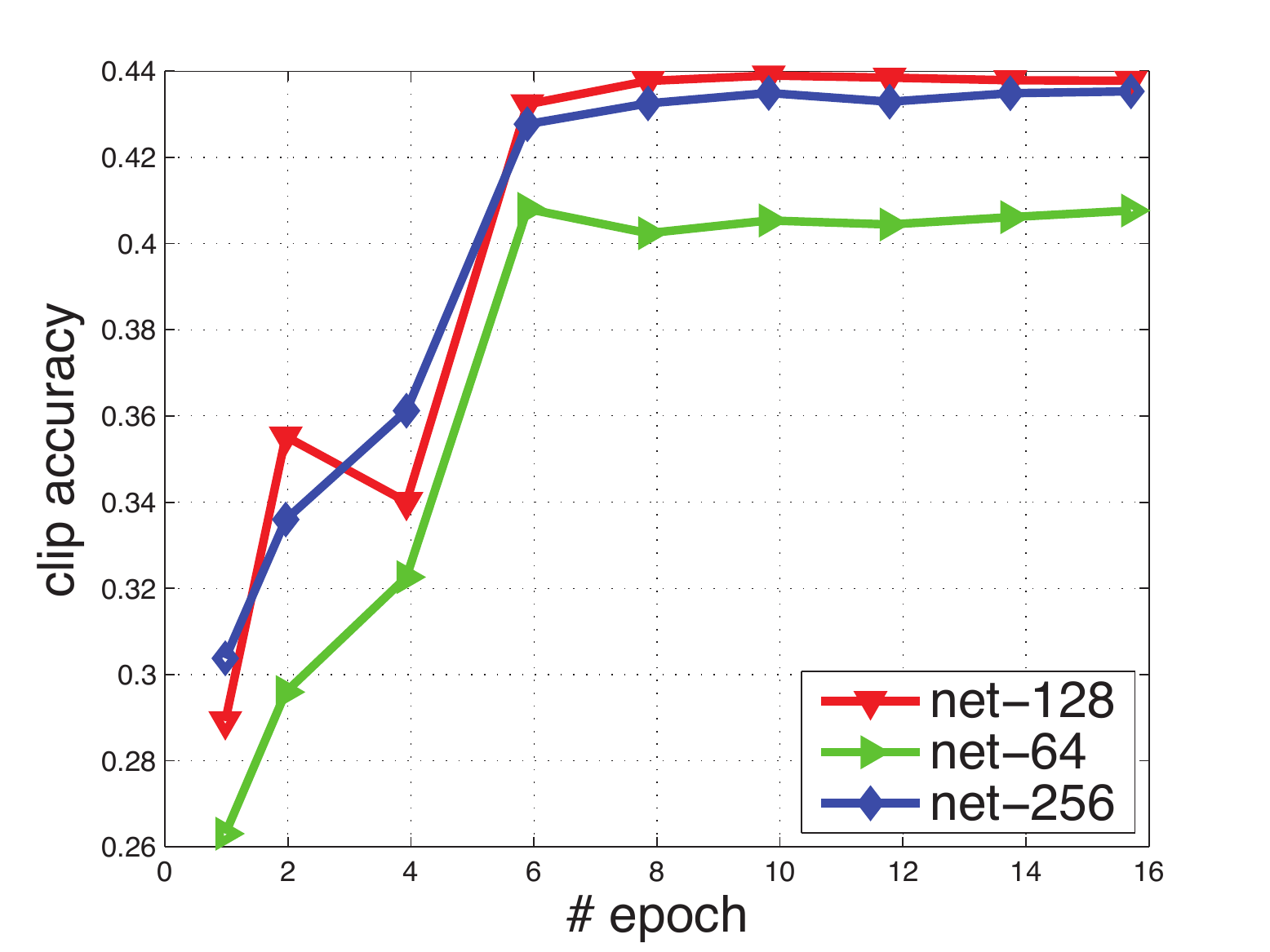}
\end{center}
   \caption{{\bf 3D ConvNets with different input resolutions.} Action recognition clip accuracy on UCF101 test split-1 of 3D ConvNets with different input resolutions.}
\label{fig:c3d_resolution_exp}
\end{figure}

As part of the architecture study, we examine the effects of input resolution on 3D ConvNets. We use the same common network setting described in section~\ref{sec:learning_features}. We fix all convolution kernels to $3 \times 3 \times 3$ and vary the input resolutions to study the effects. We experiment with $3$ different nets with input resolutions of $64 \times 64$, $128 \times 128$, and $256 \times 256$, namely {\bf net-64}, {\bf net-128}, and {\bf net-256}, respectively. Note that \emph{net-128} is equivalent to the \emph{depth-3} net in section~\ref{sec:arch_search}. Because of the difference in input resolutions, these nets have different output size at the last pooling layer, thus leading to a significant difference in terms of number of parameters. Table~\ref{tab:params} reports the numbers of parameters and the training time of these nets. Figure~\ref{fig:c3d_resolution_exp} presents the clip accuracy of these nets on UCF101 test split-1. \emph{Net-128} outperforms \emph{net-64} by $3.1\%$ and attains a comparable accuracy with \emph{net-256}. This indicates that \emph{net-128} provides a good trade-off between training time, accuracy, and memory consumption. We note that with the current GPU memory limit, one has to use model parallelism to train C3D with $256 \times 256$ input resolution.

\begin{table}[h]
\begin{center}
\begin{tabular}{|l|c|c|c|}
\hline
Net & net-64 & net-128 & net-256 \\
\hline
$\#$ of params (M) & 11.1 & 17.5 & 34.8 \\
Train time (mins/epoch) & 92 & 270 & 1186 \\
\hline
\end{tabular}
\end{center}
\caption{Number of parameters and training time comparison of 3D ConvNets with different input resolutions. Note that net-128 is equivalent to the depth-3 net in the paper.}
\label{tab:params}
\end{table}

%% file: learned_filters.tex
\subsection*{Appendix B: Visualization of C3D Learned Features}

For a better understanding of what C3D learned internally, we provide additional visualizations using deconvolution.

{\bf Decovolutions of C3D}: We randomly select 20K clips from UCF101. We group clips that fire strongly for the same feature map at a pre-selected convolution layer. We use deconvolution~\cite{ZeilerF14} to project the top activations of these clips back into image space. We visualize the gradients causing the activiation together with the corresponding cropped image sequences. Note that we did not do any fine-tuning of C3D model on UCF101.

\begin{figure*}
\begin{center}
   \includegraphics[width=0.96\linewidth]{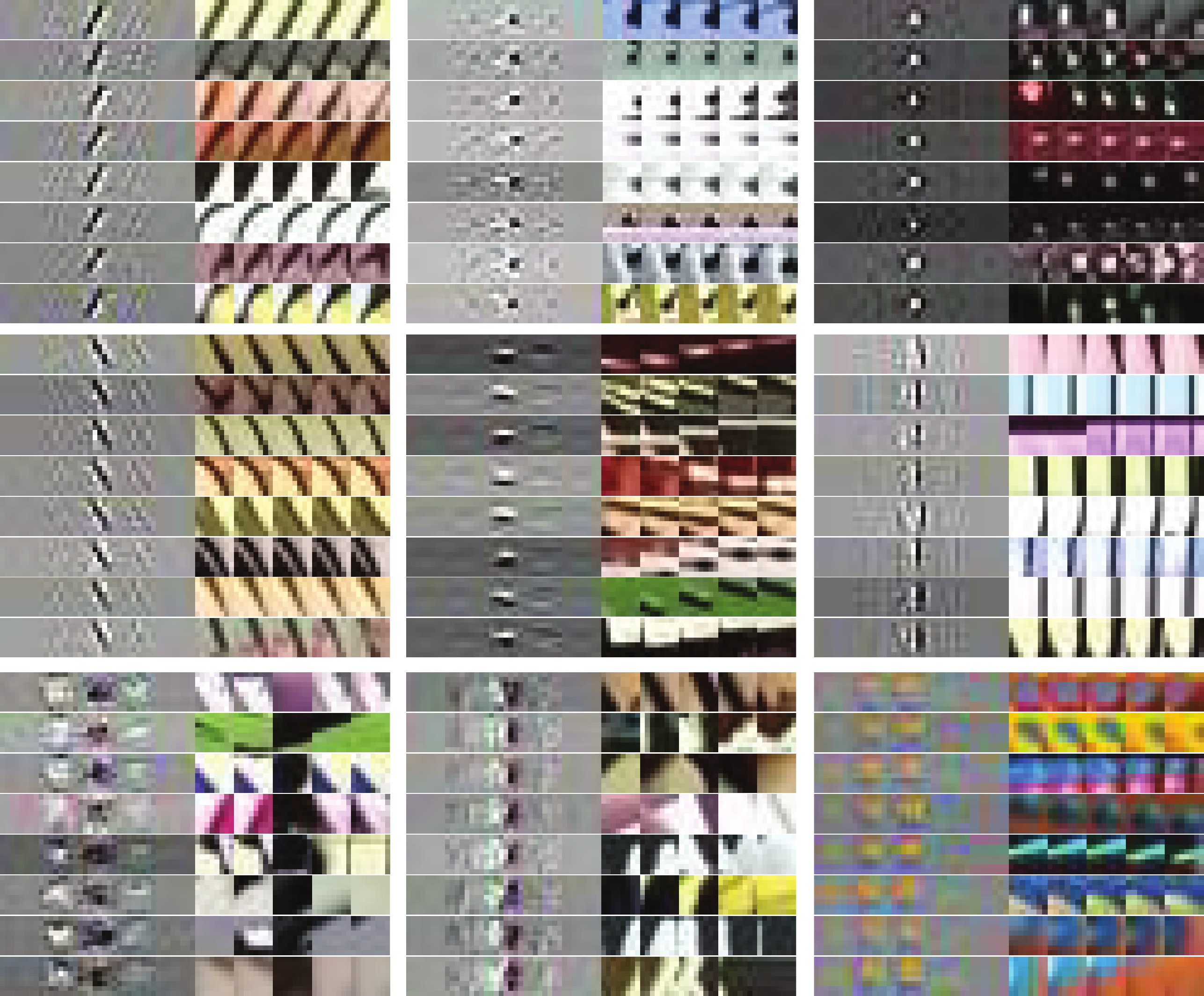}
\end{center}
   \caption{Deconvolutions of C3D \texttt{conv2a} feature maps. Each group is a C3D \texttt{conv2a} learned feature map. First two rows: the learned filters detect moving edges and blobs. The last row: the learned filters detect shot changes, edge orientation changes, and color changes. Best viewed in a color screen.}
\label{fig:conv2a}
\end{figure*}

\begin{figure*}
\begin{center}
   \includegraphics[width=0.96\linewidth]{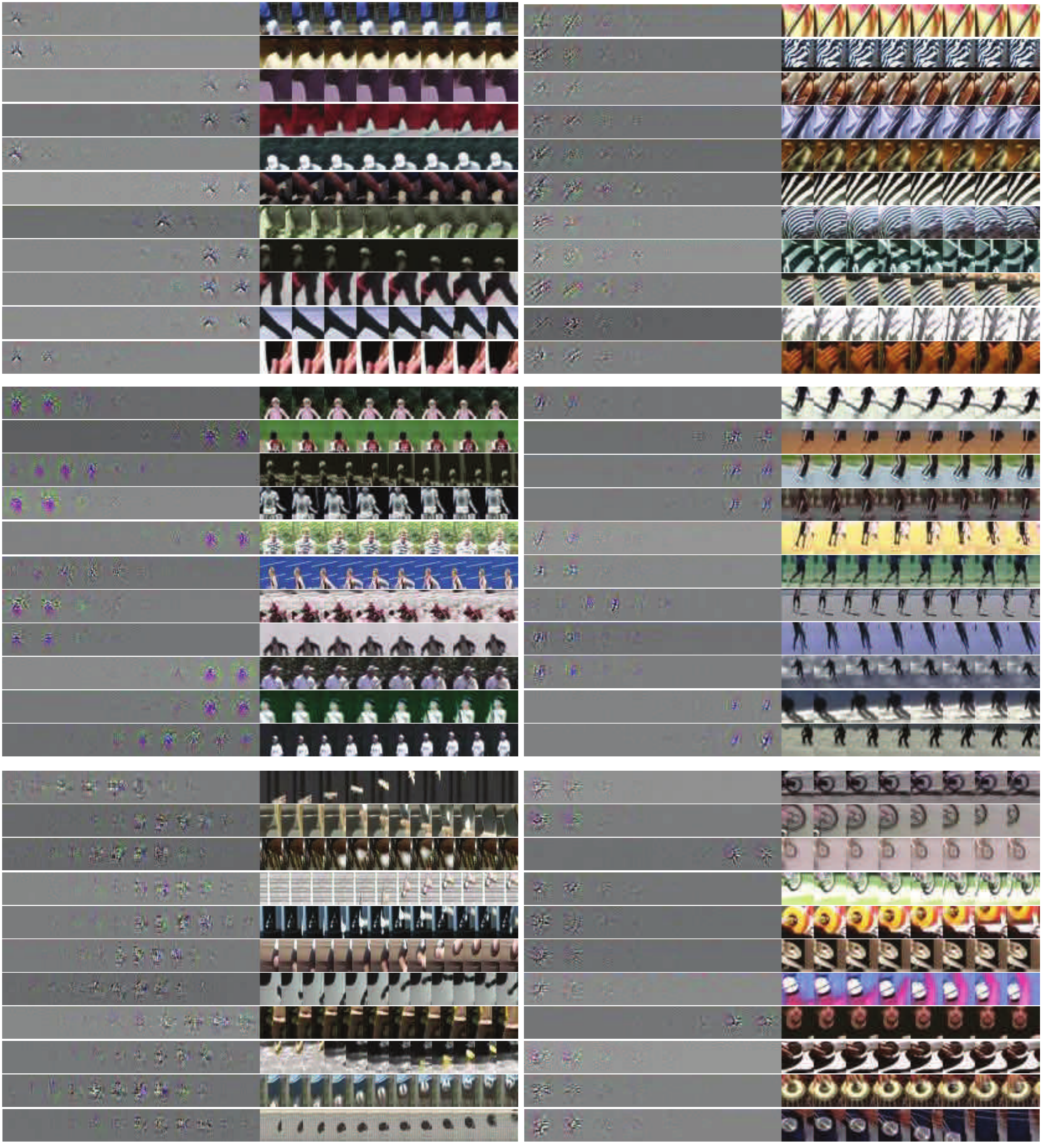}
\end{center}
   \caption{Deconvolutions of C3D \texttt{conv3b} feature maps. Each group is a C3D \texttt{conv3b} learned feature map. Upper: feature maps detect moving corners and moving textures. Middle: feature maps detect moving body parts. Lower: feature maps detect object trajectories and circular objects. Best viewed in a color screen.}
\label{fig:conv3b}
\end{figure*}

\begin{figure*}
\begin{center}
   \includegraphics[width=\linewidth]{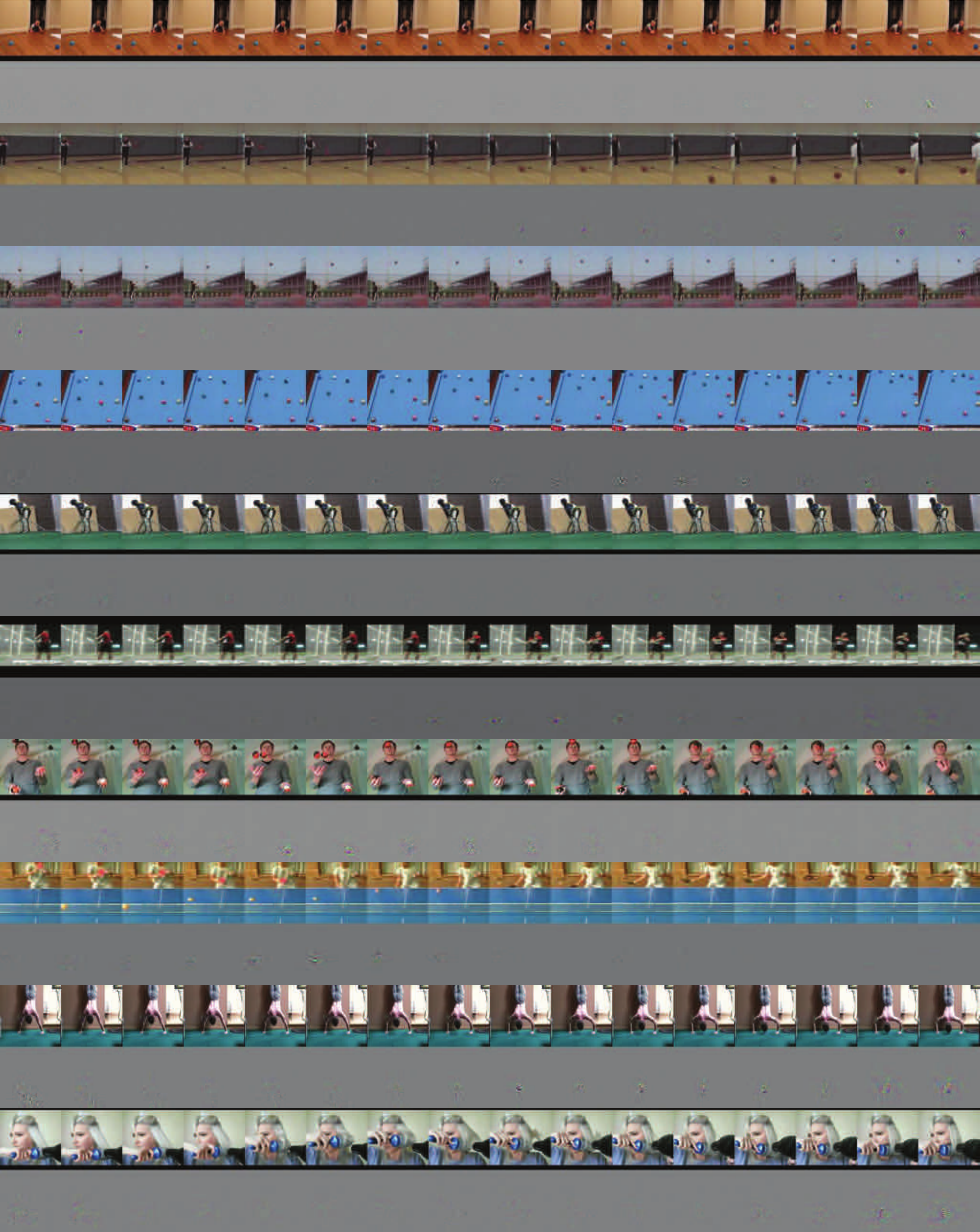}
\end{center}
   \caption{Deconvolutions of a C3D \texttt{conv5b} learned feature map which detects moving motions of circular objects. In the second last clip, it detects a moving head while in the last clip, it detects the moving hair-curler. Best viewed in a color screen.}
\label{fig:conv5b_11}
\end{figure*}

\begin{figure*}
\begin{center}
   \includegraphics[width=\linewidth]{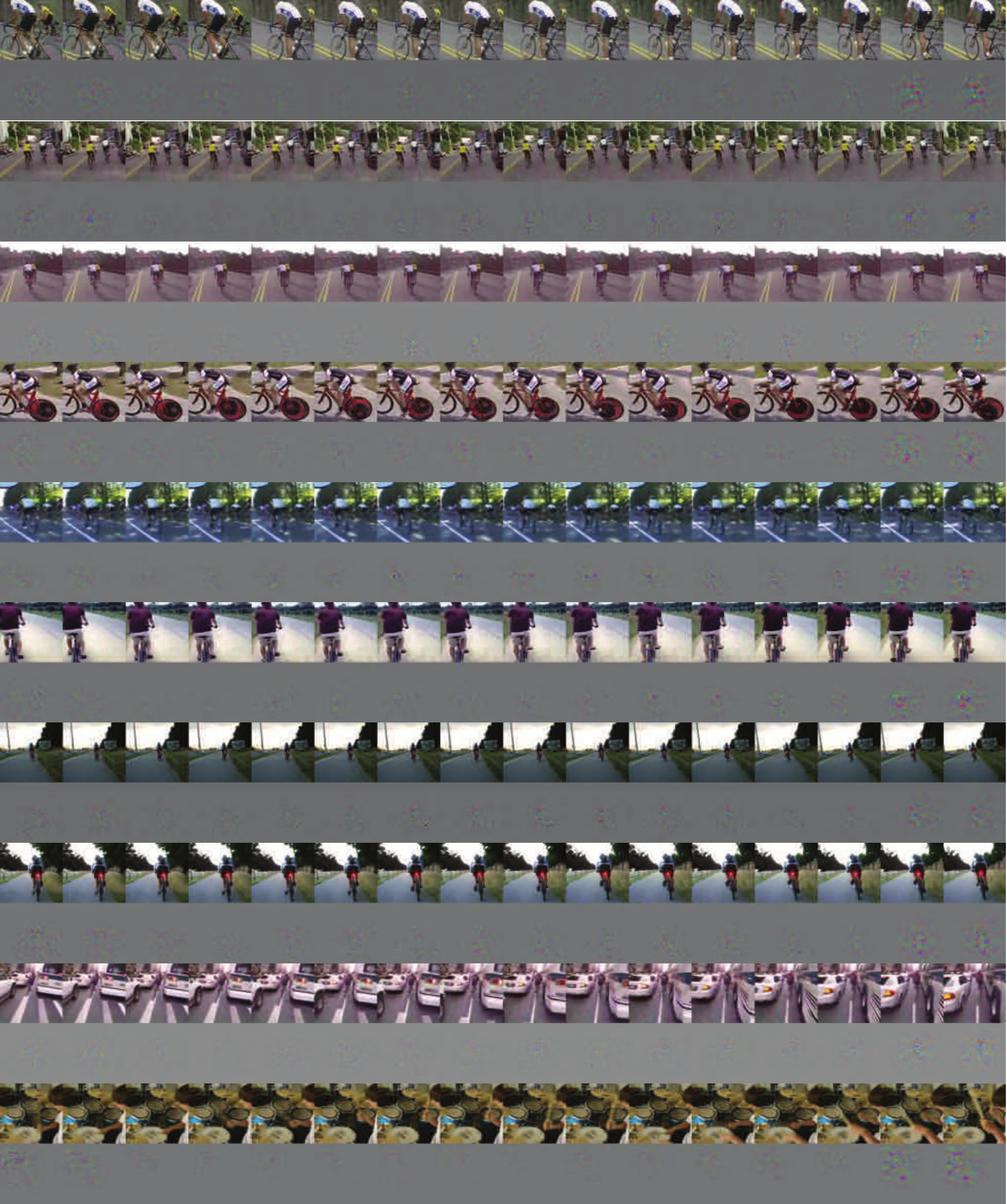}
\end{center}
   \caption{Deconvolutions of a C3D \texttt{conv5b} learned feature map which detects biking-like motions. Note that the last two clips have no biking but their motion patterns are similar to biking motions. Best viewed in a color screen.}
\label{fig:conv5b_88}
\end{figure*}

\begin{figure*}
\begin{center}
   \includegraphics[width=\linewidth]{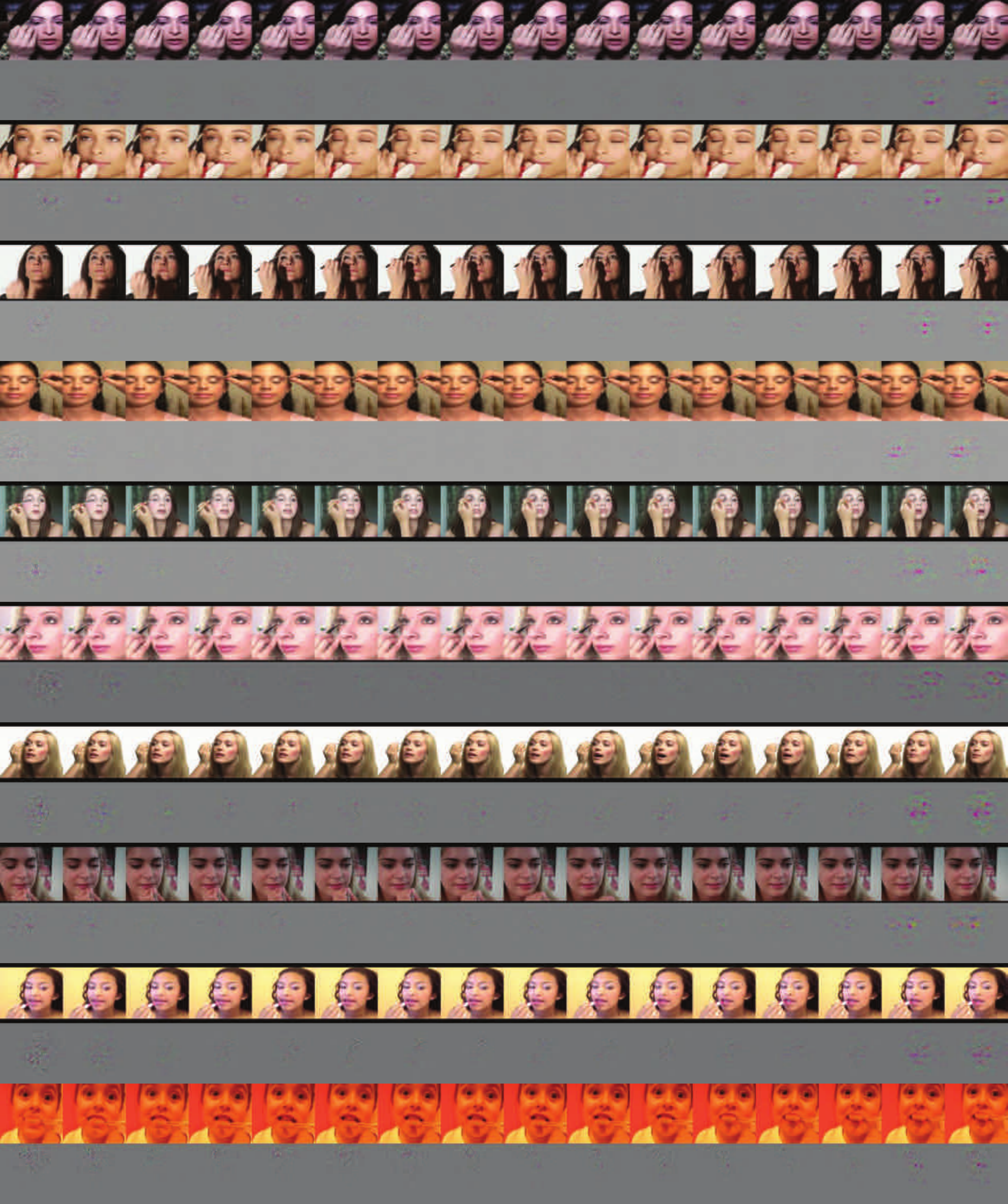}
\end{center}
   \caption{Deconvolutions of a C3D \texttt{conv5b} learned feature map which detects face-related motions: applying eye-makeup, applying lipstick, and brushing tooth. Best viewed in a color screen.}
\label{fig:conv5b_100}
\end{figure*}

\begin{figure*}
\begin{center}
   \includegraphics[width=\linewidth]{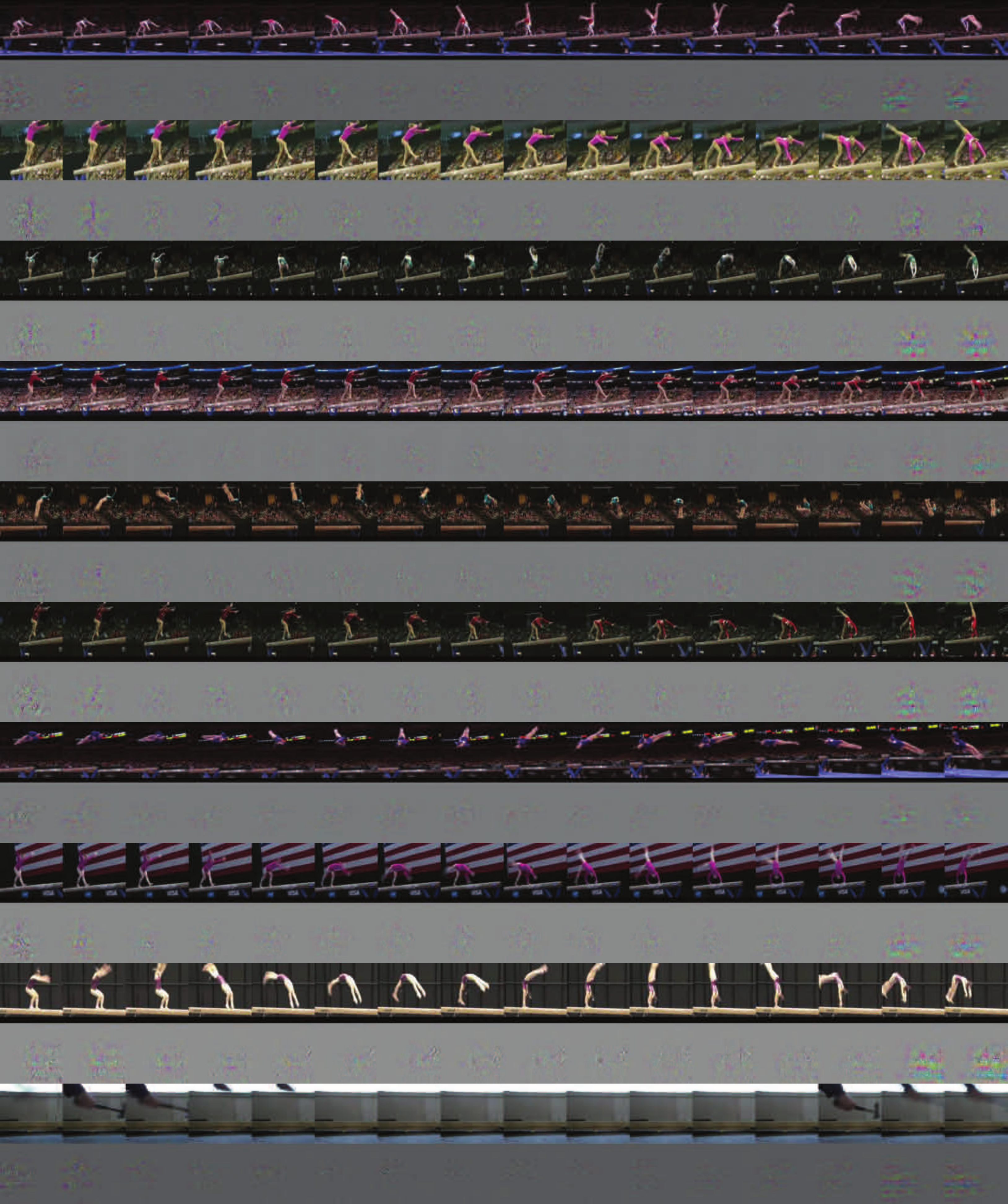}
\end{center}
   \caption{Deconvolutions of a C3D \texttt{conv5b} learned feature map which detects balance-beam-like motions. In the last clip, it detects hammering which shares similar motion patterns with balance beam. Best viewed in a color screen.}
\label{fig:conv5b_500}
\end{figure*}

\begin{figure*}
\begin{center}
   \includegraphics[width=\linewidth]{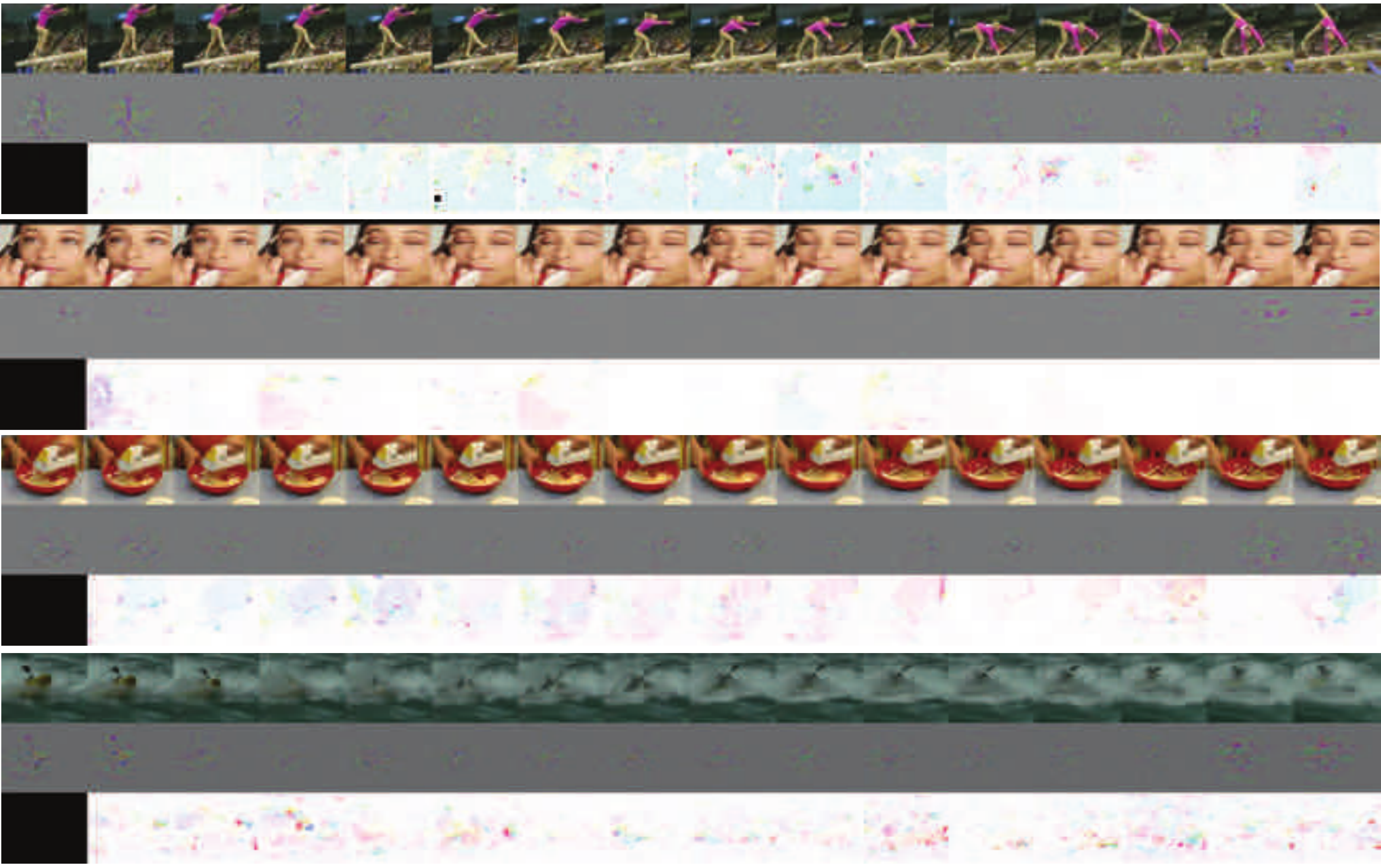}
\end{center}
   \caption{Deconvlotuions of C3D \texttt{conv5b} learned feature maps compared with optical flows. Optical flows fire at all of moving pixels while C3D just pays attention to only salient motions. Best viewed in a color screen.}
\label{fig:conv5b_flow}
\end{figure*}

Figure~\ref{fig:conv2a} and~\ref{fig:conv3b} visualize deconvolutions of C3D learned feature maps at the layers \texttt{conv2a} and \texttt{conv3b}. Visualizations of the same feature map are grouped together. For figures~\ref{fig:conv5b_11},~\ref{fig:conv5b_88},~\ref{fig:conv5b_100}, and~\ref{fig:conv5b_500}, each figure presents the deconvolutions of one learned feature map of the \texttt{conv5b} layer. Finally, figure~\ref{fig:conv5b_flow} compares the deconvolutions of several C3D \texttt{conv5b} feature maps with optical flows. As showed in the visualizations, at early convolution layer \texttt{conv2a}, C3D learns low-level motion patterns such as moving edges, blobs, short changes, edge orientation changes, or color changes. At a higher layer of \texttt{conv3b}, C3D learns bigger moving patterns of corners, textures, body parts, and trajectories. Finally, at the deepest convolution layer, \texttt{conv5b}, C3D learns more complicated motion patterns such as moving circular objects, biking-like motions.

